\documentclass[10pt,twocolumn,letterpaper]{article}

\usepackage{cvpr}              

\usepackage{graphicx}
\usepackage{amsmath}
\usepackage{amssymb}
\usepackage{booktabs}

\usepackage{bbm}

\usepackage{multirow}
\usepackage{tabularx}

\usepackage[accsupp]{axessibility}  

\DeclareMathOperator*{\argmax}{arg\,max}
\DeclareMathOperator*{\argmin}{arg\,min}

\usepackage[pagebackref,breaklinks,colorlinks]{hyperref}

\usepackage[capitalize]{cleveref}
\crefname{section}{Sec.}{Secs.}
\Crefname{section}{Section}{Sections}
\Crefname{table}{Table}{Tables}
\crefname{table}{Tab.}{Tabs.}

\def\cvprPaperID{3187} 
\def\confName{CVPR}
\def\confYear{2023}

\begin{document}

\title{UniDAformer: Unified Domain Adaptive Panoptic Segmentation Transformer via Hierarchical Mask Calibration}

\author{Jingyi Zhang\textsuperscript{\rm 1,}\thanks{Equal contribution, \{jingyi.zhang, jiaxing.huang\}@ntu.edu.sg.} \ \ Jiaxing Huang\textsuperscript{\rm 1,}$^*$ \  \  Xiaoqin Zhang\textsuperscript{\rm 2} \ \ Shijian Lu\textsuperscript{\rm 1,}\thanks{Corresponding author, shijian.lu@ntu.edu.sg.} 
\\
$^1$ S-lab, Nanyang Technological University \ \  $^2$ Wenzhou University
\\
}
\maketitle

\begin{abstract}

Domain adaptive panoptic segmentation aims to mitigate data annotation challenge by leveraging off-the-shelf annotated data in one or multiple related source domains. However, existing studies employ two separate networks for instance segmentation and semantic segmentation which lead to excessive network parameters as well as complicated and computationally intensive training and inference processes. We design UniDAformer, a unified domain adaptive panoptic segmentation transformer that is simple but can achieve domain adaptive instance segmentation and semantic segmentation simultaneously within a single network. UniDAformer introduces Hierarchical Mask Calibration (HMC) that rectifies inaccurate predictions at the level of regions, superpixels and pixels via online self-training on the fly. It has three unique features: 1) it enables unified domain adaptive panoptic adaptation; 2) it mitigates false predictions and improves domain adaptive panoptic segmentation effectively; 3) it is end-to-end trainable with a much simpler training and inference pipeline. Extensive experiments over multiple public benchmarks show that UniDAformer achieves superior domain adaptive panoptic segmentation as compared with the state-of-the-art.
\end{abstract}

\section{Introduction}

Panoptic segmentation~\cite{kirillov2019panoptic} performs instance segmentation for \textit{things} and semantic segmentation for \textit{stuff}, which assigns each image pixel with a semantic category and a unique identity simultaneously. With the advance of deep neural networks~\cite{krizhevsky2012alexnet,he2016resnet,chen2017deeplab,hoffman2016fcns,ren2015fasterrcnn,he2017mask}, panoptic segmentation~\cite{kirillov2019panoptic,li2019aunet,kirillov2019panopticfpn,xiong2019upsnet,carion2020detr,cheng2020panoptic,li2021fullypanoptic,wang2021maxdeeplab,cheng2021maskformer,cheng2021masked} has achieved very impressive performance under the supervision of plenty of densely-annotated training data. However, collecting densely-annotated panoptic data is prohibitively laborious and time-consuming~\cite{deng2009imagenet,coco,cordts2016cityscapes} which has become one major constraint along this line of research. One alternative is to leverage off-the-shelf labeled data from one or multiple \textit{source domains}. Nevertheless, the source-trained models often experience clear performance drop while applied to various \textit{target domains} that usually have different data distributions as compared with the \textit{source domains}~\cite{huang2021cvrn}.

\begin{figure}[t]
\centering
\includegraphics[width=1.0\linewidth]{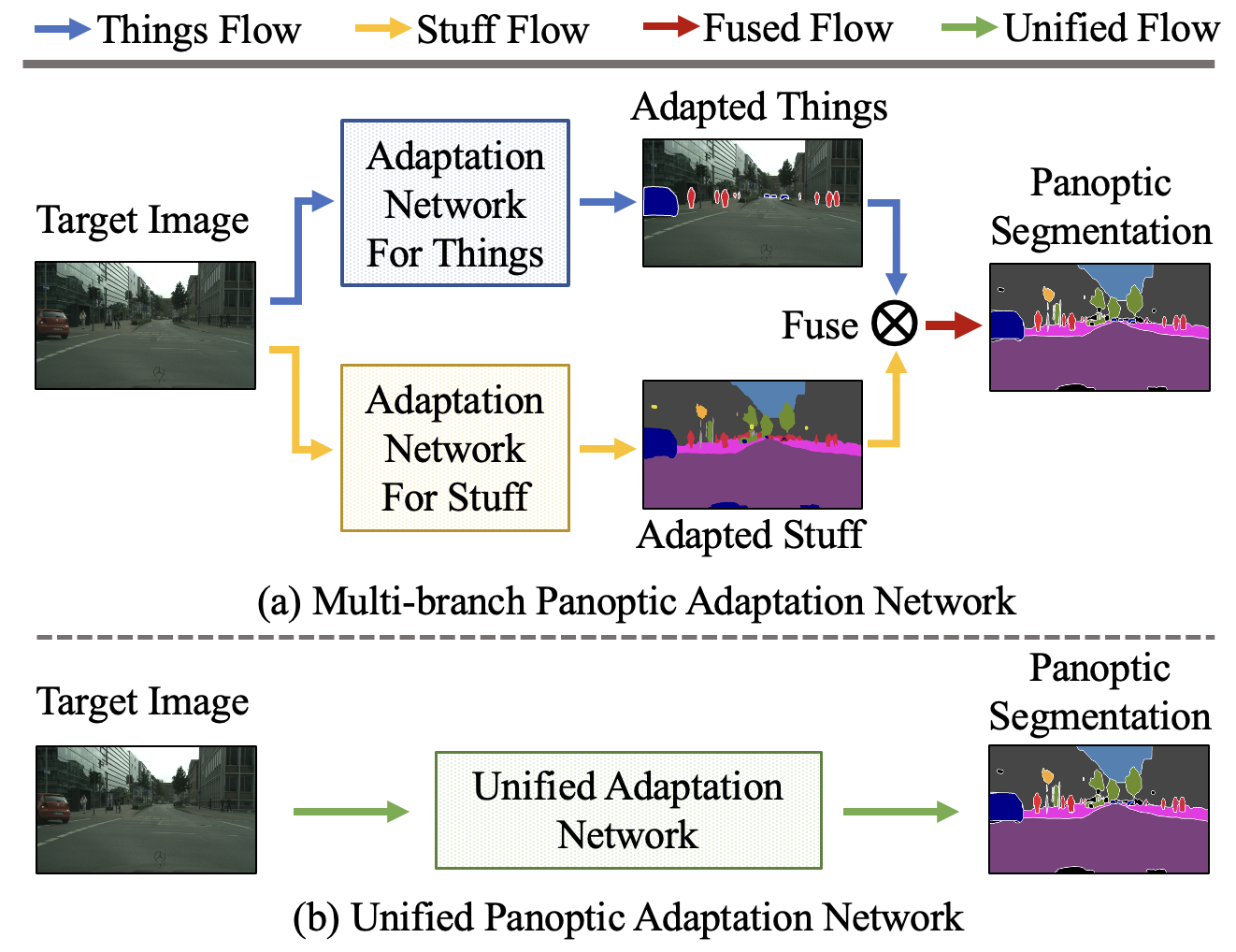}
\caption{
Existing domain adaptive panoptic segmentation~\cite{huang2021cvrn} adapts things and stuff separately with two isolated networks (for instance segmentation and semantic segmentation) and fuses their outputs to produce the final panoptic segmentation as in (a), leading to excessive network parameters as well as complicated and computationally intensive training and inference. Differently, UniDAformer employs a single unified network to jointly adapt things and stuff as in (b), which involves much less parameters and simplifies the training and inference pipeline greatly.
}
\label{fig:intro}
\end{figure}

\renewcommand\arraystretch{0.9}
\begin{table*}[t]
\centering
\resizebox{1.0\linewidth}{!}{
\begin{tabular}{c|ccc|ccc|ccc|ccc}
\toprule
&\multicolumn{3}{c|}{Multi-branch Architecture}& \multicolumn{9}{c}{Unified Architecture} \\
\cmidrule{2-13}
 & \multicolumn{3}{c|}{PSN~\cite{kirillov2019panoptic}} & \multicolumn{3}{c|}{Panoptic FCN~\cite{li2021fullypanoptic}} & \multicolumn{3}{c|}{MaskFormer~\cite{cheng2021maskformer}} & \multicolumn{3}{c}{DETR~\cite{carion2020detr}}\\
\midrule
& mSQ & \quad \ mRQ & \ \quad mPQ& \  mSQ & \ mRQ & mPQ& mSQ & mRQ & mPQ& mSQ & mRQ & mPQ
\\
\midrule
Supervised Setup & 75.5 &\quad \ 60.2& \quad 47.7& 79.7&	73.1&59.6 &79.1 &62.6&51.1 &79.1 &64.1&51.9
\\
Adaptation Setup &59.0 &\quad \ 27.8& \quad 20.1 &47.5 &19.7& 15.8 &56.6& 19.2& 16.2& 56.4&21.8&18.3 
\\
Performance Drop& -16.5 &\quad \ -32.4 & \quad -27.6 &-32.2 &-53.4 &-43.8 &-22.5 &-43.4 &-34.9 &-22.7 &-42.3 &-33.6
\\
\bottomrule
\end{tabular}
}
\caption{
Panoptic segmentation with traditional multi-branch architecture~\cite{kirillov2019panoptic} and recent unified architectures~\cite{li2021fullypanoptic,cheng2021maskformer,carion2020detr}: The \textit{Supervised Setup} trains with the Cityscapes~\cite{cordts2016cityscapes} and tests on the same dataset. The \textit{UDA Setup} trains with the SYNTHIA~\cite{ros2016synthia} and tests on Cityscapes. It can be seen that the \textit{performance drops} between the two learning setups come more from mRQ than from mSQ consistently across different architectures. In addition, such a phenomenon is more severe for unified architectures. This demonstrates a clear false prediction issue in unified domain adaptive panoptic segmentation as mRQ is computed with false positives and false negatives.
}
\label{tab:intro}
\end{table*}

Domain adaptive panoptic segmentation can mitigate the inter-domain discrepancy by aligning one or multiple labeled source domains and an unlabeled target domain~\cite{huang2021cvrn}. To the best of our knowledge, CVRN~\cite{huang2021cvrn} is the only work that tackles domain adaptive panoptic segmentation challenges by exploiting the distinct natures of instance segmentation and semantic segmentation. Specifically, CVRN introduces cross-view regularization to guide the two segmentation tasks to complement and regularize each other and achieves very impressive performance. However, CVRN relies on a multi-branch architecture that adopts a two-phase pipeline with two separate networks as illustrated in Fig.~\ref{fig:intro} (a). This sophisticated design directly doubles network parameters, slows down the training, and hinders it from being end-to-end trainable.
It is desirable to have a unified panoptic adaptation network that can effectively handle the two segmentation tasks with a single network.

We design a unified domain adaptive panoptic segmentation transformer (UniDAformer) as shown in Fig.~\ref{fig:intro} (b). Our design is based on the observation that one major issue in unified panoptic adaptation comes from a severe false prediction problem. As shown in Table~\ref{tab:intro}, most recent unified panoptic segmentation architectures~\cite{carion2020detr,li2021fullypanoptic,cheng2021maskformer} outperform traditional multi-branch architectures~\cite{kirillov2019panoptic} by large margins under the supervised setup. However, the situation inverts completely under unsupervised domain adaptation setup.
Such contradictory results are more severe for the recognition quality in mRQ. 
This shows that the panoptic quality drop of unified architecture mainly comes from False Positives (FP) and False Negatives (FN) as mRQ is computed from all predictions (True Positives, False Negatives and False Positives) while the segmentation quality in mSQ is computed with True Positives (TP) only.

In the proposed UniDAformer, we mitigate the false prediction issue by introducing Hierarchical Mask Calibration (HMC) that calibrates inaccurate predictions at the level of regions, superpixels, and pixels. With the corrected masks, UniDAformer re-trains the network via an online self-training process on the fly. Specifically, HMC treats both things and stuff predictions as masks uniformly and corrects each predicted pseudo mask hierarchically in a coarse-to-fine manner, $i.e.$, from region level that calibrates the overall category of each mask to superpixel and pixel levels that calibrate the superpixel and pixels around the boundary of each mask (which are more susceptible to prediction errors).
UniDAformer has three unique features. \textit{First}, it achieves unified panoptic adaptation by treating things and stuff as masks and adapting them uniformly. \textit{Second}, it mitigates the false prediction issue effectively by calibrating the predicted pseudo masks iteratively and progressively. 
\textit{Third}, it is end-to-end trainable with much less parameters and simpler training and inference pipeline.
Besides, HMC introduces little computation overhead and could be used as a plug-in.

The contributions of this work can be summarized in three aspects. First, we propose UniDAformer that enables concurrent domain adaptive instance segmentation and semantic segmentation within a single network. It is the first end-to-end unified domain adaptive panoptic segmentation transformer to the best our knowledge.
Second, we design Hierarchical Mask Calibration with online self-training, which allows to calibrate the predicted pseudo masks on the fly during self-training. 
Third, extensive experiments over multiple public benchmarks show that the proposed UniDAformer achieves superior segmentation accuracy and efficiency as compared with the state-of-the-art.

\section{Related Work}

\textbf{Panoptic Segmentation} is a challenging task that assigns each image pixel with a semantic category and a unique identity.
The pioneer work~\cite{kirillov2019panoptic} employs two networks for instance segmentation and semantic segmentation separately, and then combines the outputs of the two segmentation networks to acquire panoptic segmentation.
The later studies~\cite{xiong2019upsnet,kirillov2019panopticfpn,li2019aunet,carion2020detr,cheng2020panoptic,li2021fullypanoptic,wang2021maxdeeplab,cheng2021maskformer,cheng2021masked} simplify the complex pipeline by unifying the segmentation of things and stuff within single network.
For example, DETR~\cite{carion2020detr} predicts boxes around both things and stuff classes, and makes a final panoptic prediction by adding an FPN-style segmentation head.
Panoptic segmentation has achieved very impressive accuracy but requires a large amount of densely-annotated training data that are often laborious and time-consuming to collect. Domain adaptive panoptic segmentation (DAPS), which leverages off-the-shelf annotated data for mitigating the data annotation constraint, is instead largely neglected.

\begin{figure*}[t]
\centering
\includegraphics[width=1.0\linewidth]{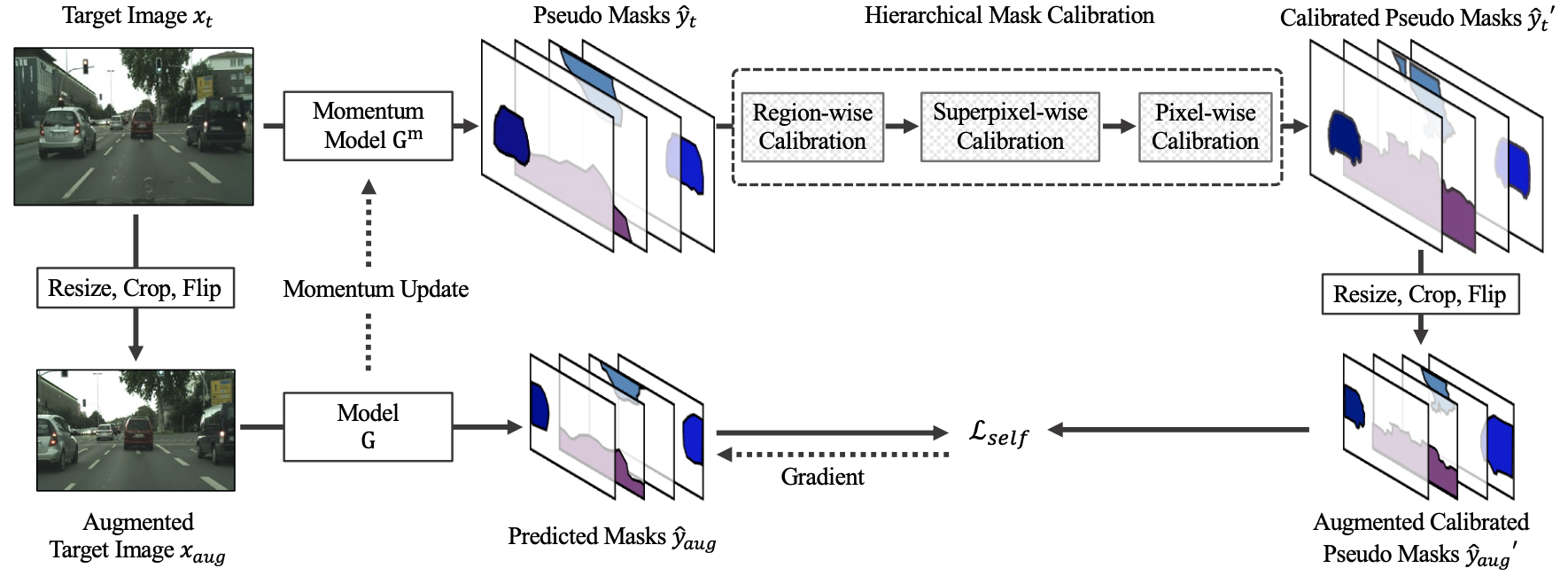}
\caption{
Overview of proposed unified domain adaptive panoptic segmentation transformer (UniDAformer):
it involves two flows, $i.e.$, a pseudo mask generation flow that calibrates pseudo masks with momentum model $G^m$, and an unsupervised training flow that optimizes model $G$ with the calibrated pseudo masks.
For pseudo mask calibration, we feed a given unlabeled target image $x_t$ into the momentum model $G^m$ to calibrate pseudo masks $\hat{y}_t$ with HMC via a coarse-to-fine manner ($i.e.$, from region level to superpixel and pixel levels).
For network optimization, we conduct simple augmentations ($i.e.$, resize, crop and flip) for $x_t$ and its calibrated pseudo masks $\hat{y}_t{'}$, and then optimize model $G$ with self-training loss $\mathcal{L}_{self}$.
}
\label{fig:stru}
\end{figure*}

\textbf{Unsupervised Domain Adaptation (UDA)} aims to exploit labeled source-domain data to learn a well-performing model on unlabeled target-domain data. In recent years, it has been studied extensively for various computer vision tasks, including image classification~\cite{ganin2015grl,saito2018maximum,pinheiro2018unsupervised,saito2019semi,zou2019confidence,sankaranarayanan2018generate,long2017deep,pinheiro2018unsupervised,lu2020stochastic,du2021cross}, instance segmentation/detection~\cite{chen2018wild,inoue2018weakly,saito2019strong,xu2020category,cai2019mtor,li2020SAP,2020coarse2fine,guan2021uncertainty,huang2021fsdr,saha2021learning,zhang2022spectral} and semantic segmentation~\cite{zhang2019category,zou2019confidence,yang2020fda,huang2020contextual,zhang2021proda,li2019bidirectional_seg,kim2020learning,pan2020unsupervised,subhani2020learning,huang2021model,huang2022category,zhang2017curriculum}. On the other hand, domain adaptive panoptic segmentation is largely neglected despite its great values in various visual tasks and practical applications. To the best of our knowledge, CVRN~\cite{huang2021cvrn} is the only work, which exploits the distinct natures of instance segmentation and semantic segmentation and introduces cross-view regularization to guide the two tasks to complement and regularize each other for panoptic adaptation.
However, CVRN achieves panoptic adaptation by using two separate adaptation networks for things and stuff respectively, which directly doubles network parameters, slows down the network, and hinders it from being end-to-end trainable.
In contrast, our proposed UniDAformer greatly simplifies training and inference pipeline by unifying the adaptation of things and stuff in a single panoptic adaptation network.

\textbf{Self-training} is a mainstream unsupervised domain adaptation technique that retrains networks with pseudo-labeled target-domain data.
Most existing self-training methods~\cite{zou2018self_seg,inoue2018weakly,iqbal2020mlsl,yu2019self-training,kim2019self,zou2019confidence,vu2019advent,lian2019constructing,huang2021model,zhang2021proda} involve an iterative retraining process for effective learning from pseudo-labeled data. In each training iteration, an offline pseudo label generation process is involved which predicts and selects pseudo labels according to their confidence. For example, ~\cite{zou2018self_seg} proposes class-balanced self-training (CBST) that globally selects the same proportion of predictions as pseudo labels for each category for overcoming class-imbalance issues.
To sidestep the cumbersome multi-round and offline training process, several studies~\cite{melas2021pixmatch,araslanov2021self} explore `online' self-training for semantic segmentation by directly enforcing pixel-wise consistency of predictions from different data augmentations. 
Differently, the proposed UniDAformer focuses on the false prediction issue in unified domain adaptive panoptic segmentation. It achieves effective `online' self-training with a Hierarchical Mask Calibration technique which allows pseudo label calibration and correction on the fly.

\section{Method}

\subsection{Task Definition}
This work focuses on domain adaptive panoptic segmentation. The training data involves a labeled source domain $\mathcal{D}_s=\left\{(x_{s}^{{i}}, y_{s}^{{i}})\right\}_{i=1}^{N_{{s}}}$ ($y_s^i$ is the panoptic annotation of sample {$x_s^i$}) and an unlabeled target domain $\mathcal{D}_t = \left\{x_{t}^{{i}}\right\}_{i=1}^{N_{{t}}}$. The goal is to learn a model $G$ from $\mathcal{D}_s$ and $\mathcal{D}_t$ that well performs in $\mathcal{D}_t $. The baseline model is trained with the source domain data $\mathcal{D}_s$ only:

\begin{equation}
\label{eq1}
\mathcal{L}_{sup}=l(G(x_s)), y_s),
\end{equation}
where $l(\cdot)$ denotes the panoptic segmentation loss that consists of a matching cost and a Hungarian loss~\cite{carion2020detr}.

\begin{figure*}[t]
\centering
\includegraphics[width=1.0\linewidth]{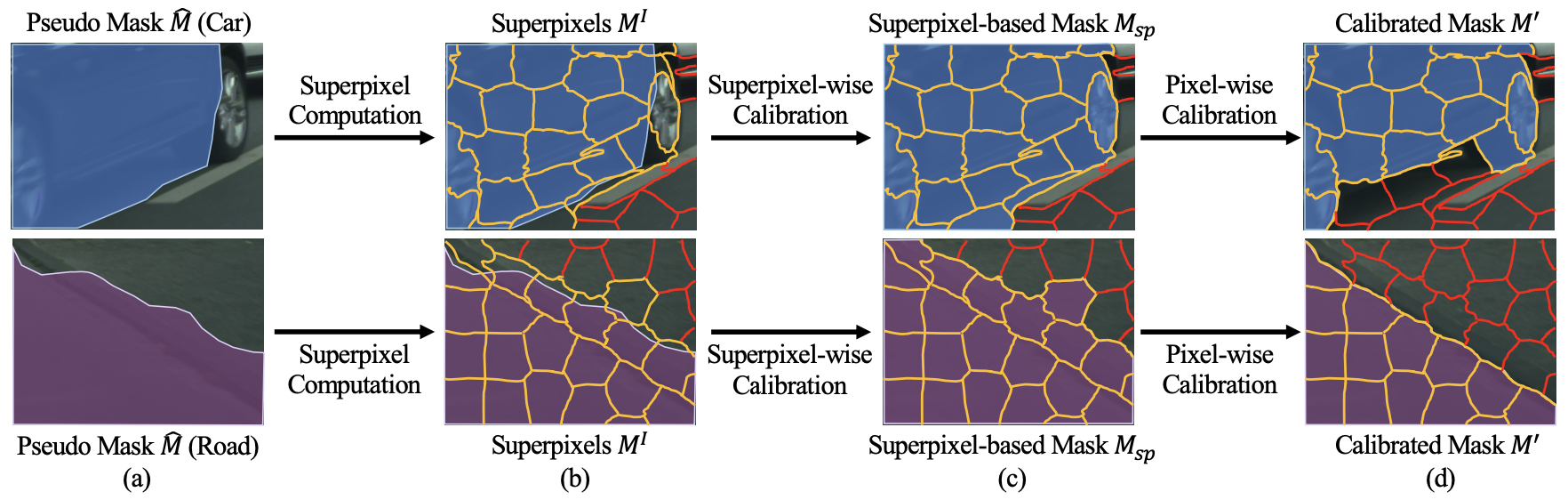}
\caption{
Overview of Hierarchical Mask Calibration:
it consists of three sub-modules, $i.e.$, Region-wise Calibration Module, Superpixel-wise Calibration Module and Pixel-wise Calibration Module. 
For simplicity, we skip visualizing Region-wise Calibration and directly present pseudo mask after region-wise calibration in (a).
In Superpixel-wise Calibration, we first compute superpixels $M^I$ and select the superpixels (marked with yellow lines as in (b)) that overlap with the pseudo mask, based on which the pseudo mask is expanded into superpixel-based mask $M_{sp}$ as in (c). In Pixel-wise Calibration, we discard the superpixels that are inconsistent with the calibrated overall category through a pixel-wise voting mechanism to form the final calibrated mask $M^{'}$ as in (d).
}
\label{fig:shape}
\end{figure*}

\subsection{UniDAformer Overview}

This subsection presents the overall framework of proposed UniDAformer, which consists of a supervised training process over the labeled source domain and an unsupervised training process over the unlabeled target domain. 
For the \textit{supervised training}, the source samples $(x_s, y_s)$ are fed to a panoptic segmentation model $G$ that is optimized via the supervised loss $\mathcal{L}_{sup}$ as defined in Eq.~\ref{eq1}.

The \textit{unsupervised training} involves two flows as illustrated in Fig.~\ref{fig:stru}. The first flow calibrates pseudo masks with the momentum model $G^m$ (the moving averaged of $G$, $i.e.$, $\theta_{G^m} \leftarrow \gamma \ \theta_{G^m} + (1 - \gamma) \theta_{G}$, and $\gamma$ is a momentum coefficient). In pseudo mask calibration, we first feed target image $x_t$ into the momentum model $G^m$ and calibrate the pseudo masks $\hat{y}_t$.
The pseudo masks $\hat{y}_t$ are then forwarded to the Hierarchical Mask Calibration (HMC) module that produces corrected pseudo masks $\hat{y}_t^{\prime}$ via coarse-to-fine calibration. The second flow optimizes $G$ with the calibrated pseudo masks. Specifically, we first apply simple data augmentations ($i.e.$, resize, crop and flip) to $x_t$ and $\hat{y}_t^{\prime}$ to obtain $x_{aug}$ and $\hat{y}_{aug}^{\prime}$. The network model $G$ is then optimized with the augmented data and the self-training loss $\mathcal{L}_{self}$ as defined in Eq.~\ref{eq9}.

\subsection{Hierarchical Mask Calibration}

One key component in the proposed UniDAformer is HMC that calibrates the predicted pseudo masks and enables effective pseudo-label retraining on the fly.
HMC treats both things and stuff predictions as masks uniformly and corrects each predicted mask hierarchically in a coarse-to-fine manner. The correction involves three consecutive stages of calibration including a Region-wise Calibration, a Superpixel-wise Calibration and a Pixel-wise Calibration as illustrated in Fig.~\ref{fig:stru}.
First, Region-wise Calibration corrects the overall category of each mask by adaptively re-weighting its category-wise probabilities.
Leveraging the feature that superpixels adhere well to the boundaries of things and stuff, Superpixel-wise Calibration then adjusts the shape of each mask by considering the boundary of the computed superpixels.
Finally, Pixel-wise Calibration introduces pixel-level categorization information and further refines the boundary of each mask with a simple pixel-wise voting mechanism.

As the proposed calibration technique works for all predicted pseudo masks ($i.e.$, things and stuff) uniformly, we take one pseudo mask $\hat{y}_t^{k}$ from $\hat{y}_t = \{ \hat{y}_t^{k} \}_{k=1}^{K}$ as an example for illustration. Each pseudo mask $\hat{y}_t^{k}$ includes a predicted category $\hat{c}_t^k = \argmax_c p^{c}$ ($p^{c} \in \{ p^{c} \}_{c=1}^{C}$ is the probability of belonging to the $c$-th category) and a predicted binary mask $\hat{M}_t^k$ of size $H \times W$.

\noindent \textbf{Region-wise Calibration} corrects the predicted category $\hat{c}_t^k$ by re-weighting its category-wise probability $p^{c}$ as following:
\begin{equation}
\label{eq2}
 {c_t^k}^{\prime} = \argmax_c (w^{(c,k)} \otimes p^{c}), 
\end{equation}
where $\otimes$ denotes the element-wise multiplication and $w^{(c,k)} \in \{w^{(c,k)} \}_{c=1}^C $ is the calibration weight of the corresponding $c$-th category probability for each pseudo mask. 

The calibration weight $w$ is calculated according to the distance between $\hat{y}_t^{k}$ and the centroids $\delta$ over feature space ($i.e.$, feature center calculated and updated in Eqs.~\ref{eq5} and~\ref{eq6}). Given the feature $f_t^k \in \mathbb{R}^{E \times H \times W}$ ($E$, $H$ and $W$ denote the feature's channel, height and width respectively) as generated by the momentum backbone, we pool the feature within the mask $\hat{M}$ into a region-wise vector $v_r^k \in \mathbb{R}^{E}$ (subscript $r$ denotes it is a region-wise vector) as follows:

\begin{equation}
\label{eq3}
  v_r^k = \text{GAP}(\hat{M}_t^k \otimes f_t^k), 
\end{equation}
where $\text{GAP}(\cdot)$ denotes the global average pooling operation.

Generally, if the region-wise vector $v_r^k$ is far from the $c$-th centroid $\delta^c$, the pseudo mask $\hat{y}_t^{k}$ should be assigned with a lower probability of belonging to the $c$-th category, and vice versa. Therefore, the calibration weight in Eq.~\ref{eq2} is defined as follows:
\begin{equation}
\label{eq4}
 w^{(c,k)} = \text{Softmax} (-||v_r^k - \delta^c ||_{1}) , 
\end{equation}
where the distance is measured using L1 distance and softmax operation is performed along the category dimension.

Here we demonstrate how we compute and update the mask centroids along the training process. The mask centroids are first initialized by all target predictions from the baseline model. For each category, the mask centroid $\delta^c$ is defined as follows:
\begin{equation}
\label{eq5}
\delta^{c} = \frac{\sum_{x_t \in \mathcal{D}_t} \sum_{k\in K} v^r_m \cdot \mathbbm{1}(\hat{c}^k = c)}{\sum_{x_t \in \mathcal{D}_t} \sum_{k\in K}\mathbbm{1}(\hat{c}^k = c) } ,
\end{equation}
where $\mathbbm{1}$ is an indicator function that returns `1' if the vector $v_m^k$ belongs to $c$-th category, and `0' otherwise.

Along training process, we update the mask centroids with the current batch of data:
\begin{equation}
\label{eq6}
 \delta^{c} \leftarrow \gamma^{\prime} \delta^{c} + (1-\gamma^{\prime}) \delta_*^c ,
\end{equation}
where $\delta_*^c$ is the mask centroid calculated with the current data and model, and $\gamma^{\prime}$ is a update coefficient for smooth centroid update.

\noindent \textbf{Superpixel-wise Calibration:}
Following region-wise calibration, we first correct the shape of the pseudo mask by exploiting superpixels that adhere well to the boundaries of things and stuff~\cite{achanta2012slic}.
To this end, we first compute a superpixel map $M^I$ which includes total $I$ superpixels $M^{(i)}$ for target image $x_t$. Then, we select the superpixels that overlap with the original mask $\hat{M}_t^k$ to form an adjusted binary mask $M_{sp}^k$ as follows:
\begin{equation}
\label{eq7}
M_{sp}^k = \bigcup_{i\in I} M^{(i)} \cdot \mathbbm{1} (A^{(i)}> 0), 
\end{equation}
where $\mathbbm{1}$ is an indicator function and we denote the overlapping area between $i$-th superpixel and the mask $\hat{M}_t^k$ as $A^{(i)}$.

The superpixel-based mask $M_{sp}^k$ adjusts the original mask $\hat{M}$ with the computed superpixels which adheres better to the edge of things or stuff, as illustrated in Figs.~\ref{fig:shape} (b) and (c).

\noindent \textbf{Pixel-wise Calibration:} 
Following superpixel-wise calibration, we further introduce pixel-level categorization information and refine the boundary of the superpixel-based mask $M_{sp}^k$ in a more precise manner. 

We design a simple pixel-wise voting mechanism to achieve the fine-grained calibration: the superpixels are discarded from the $M_{sp}^k$ if a majority of pixel-wise feature vectors within it are inconsistent with the overall category of pseudo mask $\hat{y}_t^k$ as illustrated in Fig.~\ref{fig:shape} (d).
Let $v_p^{(k,j)}$ denote a pixel-wise feature vector within superpixel $M^{(j)} \in M_{sp}$ (subscript $p$ denotes it is a pixel-wise vector), and we determine that it is inconsistent with the pseudo mask if it does not satisfy the following condition:
\begin{equation}
\label{eq8}
\argmin_c (||v_p^{(k,j)} - \delta^c ||) = {c_t^k}^{\prime},
\end{equation}
where ${c_t^k}^{\prime}$ is the corrected category of the pseudo mask $\hat{y}_t^k$. 
Such pixel-wise voting effectively suppresses the side effect of outlier pixels by enforcing that pixels within the same superpixel should share the same semantic category~\cite{achanta2012slic}. 

The final calibrated pseudo masks $\hat{y}_t^{\prime}$ for target image $x_t$ can be acquired by repeating the above-mentioned processes for all the calibrated pseudo mask ($i.e.$, $\hat{y}_t^{\prime} = \{\hat{y}_t^{k'}\}_{k=1}^{K}$, where $\hat{y}_t^{k'} = \{{c_t^k}^{\prime}, {M_t^k}^{\prime}\}$). 

\renewcommand\arraystretch{1.0}
\begin{table}[t]
\centering
\begin{footnotesize}
\resizebox{1.0\linewidth}{!}{
\begin{tabular}{cccc|ccc}
\toprule
\multicolumn{1}{c|}{{Self-train.}} &\multicolumn{1}{c|}{Region} &\multicolumn{1}{c|}{Superpixel}&\multicolumn{1}{c|}{Pixel} & mSQ & mRQ & mPQ
\\
\midrule
&&&& 56.4&21.8&18.3\\
\midrule
\multicolumn{1}{c}{\checkmark}& & &&59.5&29.9&22.6\\
\multicolumn{1}{c}{\checkmark} &\multicolumn{1}{c}{\checkmark} & & &61.2&36.9&28.7
\\
\multicolumn{1}{c}{\checkmark} & & \multicolumn{1}{c}{\checkmark} & &63.0&32.4&26.2
\\
\multicolumn{1}{c}{\checkmark} & & & \multicolumn{1}{c|}{\checkmark} &62.6 &31.8 &24.2
\\
\multicolumn{1}{c}{\checkmark} &\multicolumn{1}{c}{\checkmark} & \multicolumn{1}{c}{\checkmark}& & 63.4&39.9&30.9
\\
\multicolumn{1}{c}{\checkmark} &\multicolumn{1}{c}{\checkmark} & &\multicolumn{1}{c|}{\checkmark} &63.2&38.9&30.1
\\
\multicolumn{1}{c}{\checkmark} & &\multicolumn{1}{c}{\checkmark} &\multicolumn{1}{c|}{\checkmark} &64.3&32.7&26.9
\\
\multicolumn{1}{c}{\checkmark} &\multicolumn{1}{c}{\checkmark} &\multicolumn{1}{c}{\checkmark} &\multicolumn{1}{c|}{\checkmark} &\textbf{64.7}&\textbf{42.2}&\textbf{33.0}\\
\bottomrule
\end{tabular}
}
\end{footnotesize}
\caption{
Ablation study of the proposed Hierarchical Mask Calibration technique over task SYNTHIA $\rightarrow$ Cityscapes, where `Region', `Superpixel' and `Pixel' stand for region-wise calibration, superpixel-wise calibration and pixel-wise calibration, respectively.
}
\label{ablation_1}
\end{table}

\renewcommand\arraystretch{1.0}
\begin{table*}[t]
\centering
\resizebox{1.0\linewidth}{!}{
\begin{tabular}{c|ccccccccccccccccccc}
 \toprule
 \multicolumn{20}{c}{\textbf{SYNTHIA $\rightarrow$ Cityscapes Panoptic Segmentation}} \\
 \midrule
 Methods &road &side. &build.&wall &fence &pole &light&sign &vege. &sky&pers.&rider&car&bus&mot.&bike&mSQ&mRQ&mPQ\\
 \midrule
Baseline~\cite{carion2020detr} &33.9 &7.3 &45.6 &0.0 &0.0 &2.9&5.2&7.6&65.0&57.1&19.2&5.4&22.3&14.9&1.3&4.8& 56.4&21.8&18.3
\\
 DAF~\cite{chen2018wild} & 34.6&7.5&53.8&0.0&0.0&2.1&3.1&2.6&72.8&67.5&21.3&9.4&28.2&22.7&4.4&4.4&59.0&28.3&20.9
\\
 FDA~\cite{yang2020fda} & 29.2&5.8&63.6&0.1&0.0&4.9&4.0&4.8&73.9&62.5 &24.5&11.4&32.1&27.1&5.2&8.0&59.1&30.3&22.3
\\
 CRST~\cite{zou2019confidence}&46.8&11.8&56.8&0.9&0.0&4.0&2.6&3.5&70.9&64.3&20.7&11.8&32.9&32.2&6.2&7.9&62.5&31.9&23.3
 \\
SVMin~\cite{guan2021scale}&48.0&11.8&56.9&0.5&0.0&3.9&3.8&4.4&72.5&68.9& 26.4&15.0&35.3&25.8&6.3&7.6&63.3&32.6&24.2
\\
 AdvEnt~\cite{vu2019advent} & 55.9&14.4&64.0&0.0&0.0&4.6&3.3&2.7&75.5&72.3&24.9&9.7&33.8&26.7&5.2&7.1 &60.2&33.0&25.0
\\
\midrule
 CVRN~\cite{huang2021cvrn} & 66.2&19.4&\textbf{72.5}&\textbf{2.1}&0.0&3.8&6.5&4.4&79.7&75.1&26.5&11.5&36.6&34.1&7.1&8.2&61.4&35.9&27.9
\\
\textbf{UniDAformer} &\textbf{73.7}&\textbf{26.5}&71.9&1.0&0.0&\textbf{7.6}&\textbf{9.9}&\textbf{12.4}&\textbf{81.4}&\textbf{77.4}&\textbf{27.4}&\textbf{23.1}&\textbf{47.0}&\textbf{40.9}&\textbf{12.6}&\textbf{15.4}&\textbf{64.7}&\textbf{42.2}&\textbf{33.0}
\\
\bottomrule
\end{tabular}
}
\caption{Experiments with unified panoptic segmentation architecture~\cite{carion2020detr} over task SYNTHIA $\rightarrow$ Cityscapes. PQ is computed for each category. Mean SQ (mSQ), mean RQ (mSQ), mean PQ (mPQ) are computed over all categories.}
\label{tab:syn2city}
\end{table*}

\renewcommand\arraystretch{1.1}
\begin{table*}[t]
\centering
\resizebox{1.0\linewidth}{!}{
\begin{tabular}{c|ccccccccccccccccccc}
 \toprule
 \multicolumn{20}{c}{\textbf{Cityscapes $\rightarrow$ Foggy Cityscapes Panoptic Segmentation}} \\
 \midrule
 Methods &road &side. &build.&wall &fence &pole &light&sign &vege. &sky&pers.&rider&car&bus&mot.&bike&mSQ&mRQ&mPQ \\
 \midrule
 Baseline~\cite{carion2020detr} &92.5&48.9&60.6&6.0&10.7&5.3&9.9&23.6&49.7&55.6&22.3&15.4&38.5&23.7&1.6&2.8&70.0&38.6&29.2
\\
 DAF~\cite{chen2018wild} & 94.0&\textbf{54.5}&57.7&6.7&10.0&7.0&6.6&25.5&44.6&\textbf{59.1}&26.7&16.7&42.2&36.6&4.5&16.9&70.6&41.7&31.8
\\
 FDA~\cite{yang2020fda} & 93.8& 53.1& 62.2& 8.2& 13.4& 7.3& 7.6& \textbf{28.9}& 50.8& 49.7&25.0& 22.6& 42.9& 36.3& 10.3& 15.2& 71.4&43.5& 33.0\\
 AdvEnt~\cite{vu2019advent} & 93.8&52.7&56.3&5.7&13.5&\textbf{10.0}&\textbf{10.9}&27.7&40.7&57.9&27.8&29.4&44.7&28.6&11.6&20.8&72.3&43.7&33.3
\\
 CRST~\cite{zou2019confidence}&91.8&49.7&66.1&6.4&14.5&5.2&8.6&21.5&56.3&50.7&30.5&30.7&46.3&34.2&11.7&22.1&72.2&44.9&34.1
 
\\
 SVMin~\cite{guan2021scale}&
  93.4& 53.4& 62.2& \textbf{12.3}& 15.5& 7.0& 8.5& 18.0& 54.3& 57.1& 31.2& 29.6& 45.2& 35.6& 11.5& 22.7&72.4&45.5& 34.8\\
  \midrule
 CVRN~\cite{huang2021cvrn} &93.6&52.3&\textbf{65.3}&7.5&\textbf{15.9}&5.2&7.4&22.3&\textbf{57.8}&48.7&32.9&30.9&49.6&38.9&18.0&25.2&72.7&46.7&35.7
\\
 \textbf{UniDAformer} &
\textbf{93.9}& 53.1& 63.9& 8.7& 14.0& 3.8&10.0& 26.0& 53.5& 49.6 &\textbf{38.0}& \textbf{35.4}& \textbf{57.5}& \textbf{44.2}& \textbf{28.9}& \textbf{29.8}&\textbf{72.9}&\textbf{49.5}&\textbf{37.6}
\\
\bottomrule
\end{tabular}
}
\caption{Experiments with unified panoptic segmentation architecture~\cite{carion2020detr} over task Cityscapes $\rightarrow$ Foggy cityscapes. PQ is computed for each category. Mean SQ (mSQ), mean RQ (mSQ), mean PQ (mPQ) are computed over all categories.}
\label{tab:city2foggy}
\end{table*}

\subsection{Network Optimization}

With the calibrated pseudo masks $\hat{y}_t^{\prime}$, the self-training loss $\mathcal{L}_{self}$ can be formulated as follows:
\begin{equation}
\label{eq9}
\mathcal{L}_{self} = l(G(x_{aug}), \hat{y}_{aug}^{\prime}),
\end{equation}
where $l(\cdot)$ denotes the panoptic segmentation loss that consists of a matching cost
and a Hungarian loss~\cite{carion2020detr}. $\hat{y}_{aug}^{\prime}$ and $x_{aug}$ are the simple augmentations ($i.e.$, resize, crop and flip) of $\hat{y}_t^{\prime}$ and $x_t$, respectively.

The overall training objective is defined by minimizing the supervised loss $\mathcal{L}_{sup}$ and the unsupervised loss $\mathcal{L}_{self}$:
\begin{equation}
\label{eq10}
\argmin_G \mathcal{L}_{sup} + \mathcal{L}_{self}.
\end{equation}

\section{Experiment}
This section presents experiments including datasets, evaluation metric, ablation studies, comparisons with the state-of-the-art and discussions. 
Due to the space limit, the implementation details are provided in the appendix.

\subsection{Datasets}
\label{sec:details}
We evaluate UniDAformer over three widely used domain adaptation tasks with four datasets:

\noindent 1) SYNTHIA~\cite{ros2016synthia} $\rightarrow$ Cityscapes~\cite{cordts2016cityscapes} which aims for domain adaptation from synthetic images to real-world images. The training set in SYNTHIA are adopted as source domain and the training set in Cityscapes are considered as target domain. The evaluation is performed on the validation set of Cityscapes. 

\noindent 2) Cityscapes~\cite{cordts2016cityscapes} $\rightarrow$ Foggy cityscapes~\cite{sakaridis2018foggy} which aims for domain adaptation across different weather conditions, where Cityscapes is used as source domain and Foggy Cityscapes is considered as target domain. 
The adaptation performance is evaluated over the validation set of Foggy Cityscapes.

\noindent 3) VIPER~\cite{richter2017viper} $\rightarrow$ Cityscapes~\cite{cordts2016cityscapes} which aims for domain adaptation from synthetic images to real-world images. We adopt the training set of VIPER as source domain and the training set in Cityscapes as target domain. The evaluation is performed on the validation set of Cityscapes.

In evaluations, we adopt three panoptic segmentation metrics~\cite{kirillov2019panoptic} including segmentation quality (SQ), recognition quality (RQ) and panoptic quality (PQ) as in~\cite{kirillov2019panoptic,kirillov2019panopticfpn,li2021fullypanoptic,huang2021cvrn}. For each category, 
PQ can be computed as the multiplication of the corresponding SQ term and and RQ term as follows:
\begin{equation}
\label{pq}
 \small{\text{PQ}} = \underbrace{\frac{\sum_{(p, g) \in TP} \text{IoU}(p, g)}{\vphantom{\frac{1}{2}}|TP|}}_{\text{segmentation quality (SQ)}} \times \underbrace{\frac{|TP|}{|TP| + \frac{1}{2} |FP| + \frac{1}{2} |FN|}}_{\text{recognition quality (RQ) }} \,,
\end{equation}
where $g$ is the ground truth segment and $p$ is the matched prediction. TP, FP and FN denote true positives, false positives and false negatives, respectively. IoU is the insertion over union metric~\cite{everingham2015pascal} which is widely used in semantic segmentation evaluations. With the above definitions, RQ captures the proportion of TP in all predictions, SQ captures the segmentation quality within TP while PQ integrates PQ and SQ and captures the overall panoptic segmentation quality.

\renewcommand\arraystretch{.9}
\begin{table*}[t]
\centering
\resizebox{1.0\linewidth}{!}{
\begin{tabular}{c|cccccccccccccccc}
 \toprule
 \multicolumn{17}{c}{\textbf{VIPER $\rightarrow$ Cityscapes Panoptic Segmentation}} \\
 \midrule
 Methods &road &side. &build. &fence  &light&sign &vege. &sky&pers.&car&bus&mot.&bike&mSQ&mRQ&mPQ \\
 \midrule
 Baseline~\cite{carion2020detr} & 25.2& 5.7& 35.1& 0.0& 5.9&3.9& 75.3& 68.7& 21.2& 39.7& 21.4& 11.4& 0.0&59.7&32.0&24.1\\
 DAF~\cite{chen2018wild} &56.6& 7.3& 41.0& 0.0& 3.5& 2.7& 76.4& 70.2& 19.0& 34.3& 14.2&6.3& 0.0& 61.1&   33.3&  25.5
\\
 FDA~\cite{yang2020fda} & 50.0& 7.6& 59.4& 0.0& 6.2& 6.1& 73.3& 65.9& 19.4& 38.2& 15.5& 8.1& 0.0& 61.0 &  35.2& 26.9\\
 AdvEnt~\cite{vu2019advent}&52.6& 10.8& 51.0& 0.0& 2.0& 4.8& 73.9& 70.1& 15.9& 38.2& 19.9& 12.4& 0.0 &61.2  & 35.4& 27.0 
\\
 CRST~\cite{zou2019confidence}& 68.7&9.1& 54.4& 0.0& 2.4& 2.7& 76.3& 69.9& 21.2& 34.0& \textbf{21.9}& 7.7& 0.0& 61.0&36.5& 28.3
\\
 SVMin~\cite{guan2021scale}& \textbf{87.6}& 14.2& 70.7& 0.0& 4.1& 6.3& 74.4& 70.0& 16.9& 32.5&2.4& 11.0& 1.2 &61.3& 37.5&29.9\\
 \midrule
 CVRN~\cite{huang2021cvrn}& 75.1& 18.8& 59.9& 0.0&\textbf{ 9.1}& 6.5& 76.8& 71.1& 22.3& 37.0& 15.5& 8.6& \textbf{3.8}& 66.4&40.2&31.1\\
 \textbf{UniDAformer} &
87.1& \textbf{22.1}& \textbf{71.1}& 0.0& 8.2& \textbf{8.6}& \textbf{78.3}& \textbf{71.8}& \textbf{25.4}& \textbf{46.8}& 13.7& \textbf{12.8}& 2.8&  \textbf{68.9}&   \textbf{43.0} &\textbf{34.5}
\\
\bottomrule
\end{tabular}
}
\caption{Experiments with unified panoptic segmentation architecture~\cite{carion2020detr} over task VIPER $\rightarrow$ Cityscapes. PQ is computed for each category. Mean SQ (mSQ), mean RQ (mSQ), mean PQ (mPQ) are computed over all categories.}
\label{tab:viper2city}
\end{table*}

\renewcommand\arraystretch{1.1}
\begin{table*}[t]
\centering
\begin{footnotesize}
\resizebox{1.0\linewidth}{!}{
\begin{tabular}{c|ccccccccccccccccccc}
 \toprule
 \multicolumn{20}{c}{\textbf{SYNTHIA $\rightarrow$ Cityscapes Panoptic Segmentation}} \\
 \midrule
 Methods &road &side. &build.&wall &fence &pole &light&sign &vege. &sky&pers.&rider&car&bus&mot.&bike&mSQ&mRQ&mPQ \\
\midrule
PSN~\cite{kirillov2019panoptic}&32.3& 5.1& 58.5& 0.9& 0.0& 0.9& 0.0& 4.6& 61.7& 61.3& 27.6& 9.5 &32.8 &22.6 &1.0 &2.7& 59.0& 27.8& 20.1\\
FDA~\cite{yang2020fda}&79.0 &22.0 &61.8 &1.1 &0.0 &5.6 &5.5 &9.5 &51.6 &70.7 &23.4 &16.3 &{34.1} &31.0 &5.2 &8.8 &65.0 &35.5 &26.6 \\
CRST~\cite{zou2019confidence} &75.4 &19.0 &70.8 &1.4 &0.0 &7.3 &0.0 &5.2 &74.1 &69.2 &23.7 &{19.9} &33.4 &26.6 &2.4 &4.8 &60.3 &35.6 &27.1 \\
AdvEnt~\cite{vu2019advent}  &{87.1} &32.4 &69.7 &1.1 &0.0 &3.8 &0.7 &2.3 &71.7 &72.0 &{28.2} &17.7 &31.0 &21.1 &6.3 &4.9 &65.6 &36.3 &28.1 \\
\midrule
CVRN~\cite{huang2021cvrn}&86.6& 33.8& \textbf{74.6}& \textbf{3.4}& 0.0& \textbf{10.0}& 5.7& \textbf{13.5}& \textbf{80.3}& \textbf{76.3}& 26.0& 18.0& 34.1& \textbf{37.4}& 7.3& 6.2& 66.6& 40.9& 32.1
\\
\textbf{UniDAformer}& \textbf{87.7}& \textbf{34.0}& 73.2& 1.3& 0.0& 8.1& \textbf{9.9}& 6.7& 78.2& 74.0& \textbf{37.6}& \textbf{25.3}& \textbf{40.7}& \textbf{37.4}& \textbf{15.0}&\textbf{18.8}& \textbf{66.9}& \textbf{44.3}& \textbf{34.2}
\\
\bottomrule
\end{tabular}
}
\end{footnotesize}
\caption{Experiments with multi-branch panoptic segmentation architecture~\cite{kirillov2019panoptic} over task SYNTHIA $\rightarrow$ Cityscapes. Mean SQ (mSQ), mean RQ (mSQ), mean PQ (mPQ) are computed over all categories.}
\label{dis:cvrn}
\end{table*}

\subsection{Ablation Studies}
The core of UniDAformer is Hierarchical Mask Calibration that consists of a Region-wise Calibration, a Superpixel-wise Calibration and a Pixel-wise Calibration. We first study the three calibration modules to examine how they contribute to the overall domain adaptive panoptic segmentation.

Table~\ref{ablation_1} shows experimental results over task SYNTHIA $\rightarrow$ Cityscapes. It can be seen that the baseline in the 1st Row (trained with the labeled source data only) does not perform well due to domain shifts. Including self-training over unlabeled target data in the 2nd Row improves the baseline from 18.3 to 22.6 in mPQ. On top of the self-training, including any of the three calibration modules improves the segmentation consistently as shown in Rows 3-5. Specifically, region-wise calibration improves mRQ more (15.1 above the baseline) than the other two calibration modules (10.6 and 10.0), showing that region-wise calibration suppresses false predictions effectively by calibrating the overall category of each mask. On the other hand, superpixel-wise and pixel-wise calibrations improve mSQ more than region-wise calibration (6.6 and 6.2 vs 4.8), showing that superpixel-wise and pixel-wise calibrations focus on refining the boundary of each mask.

The three calibration modules correct pseudo masks from different levels which complement each in domain adaptive panoptic segmentation. We can observe that combining any two modules further improves mSQ, mRQ and mPQ consistently as shown in Rows 6-8, and combining all three achieves the best mSQ, mRQ and mPQ. Such experimental results are well aligned with the motivation and design of the proposed hierarchical mask calibration.

\begin{figure*}[t]
\centering
\includegraphics[width=.88\linewidth]{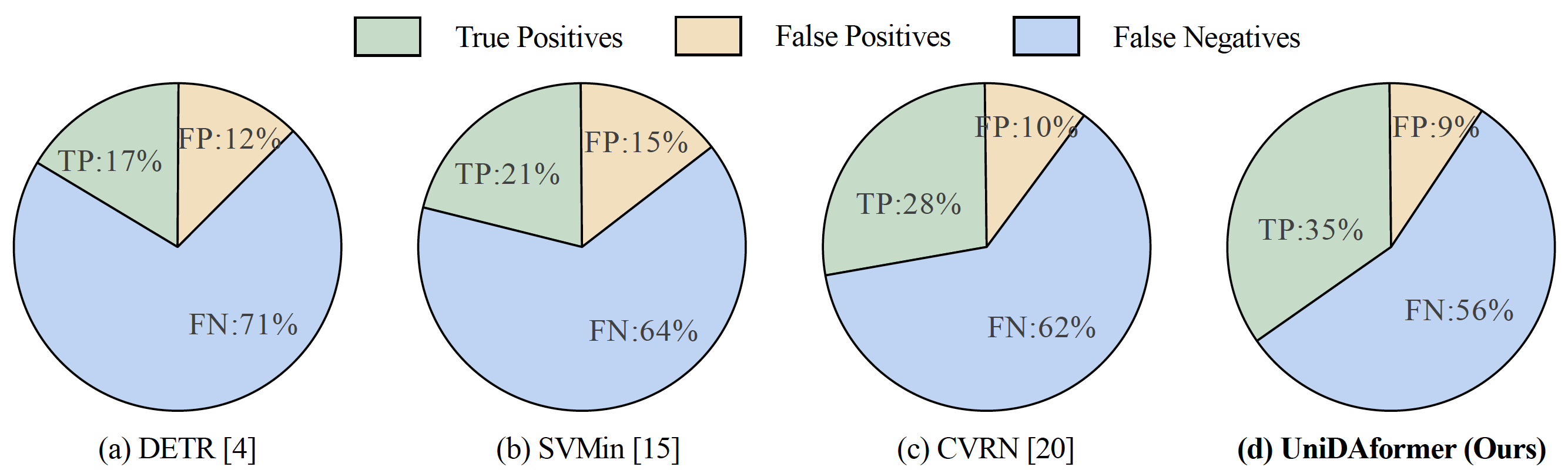}
\caption{Prediction quality analysis over task SYNTHIA $\rightarrow$ Cityscapes.
}
\label{fig:tp}
\end{figure*}

\subsection{Comparisons with the State-of-the-art}

Due to the lack of prior studies on unified domain adaptive panoptic segmentation, we conduct two sets of experiments to benchmark UniDAformer with the state-of-the-art.

In the first set of experiments, we benchmark UniDAformer over the unified panoptic segmentation architecture ($i.e.$, DETR~\cite{carion2020detr}) by reproducing the state-of-the-art~\cite{huang2021cvrn} with DETR. Specifically, we re-implement the cross-style regularization (one of two cross-view designs) in DETR to reproduce CVRN (cross-task regularization relies on multi-branch architecture and cannot work in the unified architecture). Following~\cite{huang2021cvrn}, we also reproduce several domain adaptive methods by directly implementing their adaptation module in DETR. We perform comparisons over three domain adaptive panoptic segmentation tasks as shown in Tables~\ref{tab:syn2city}-\ref{tab:viper2city}. It can be seen that UniDAformer improves the baseline~\cite{carion2020detr} by large margins (8.3, 20.4 and 14.7 in mSQ, mRQ and mPQ) and it also outperforms the state-of-the-art clearly for SYNTHIA $\rightarrow$ Cityscapes. In particular, UniDAformer improves more in mRQ as compared with the state-of-the-art, indicating that it corrects more false predictions effectively. Similar experimental results are observed on the other two tasks as shown in Tables~\ref{tab:city2foggy} and~\ref{tab:viper2city}.

In the second set of experiments, we benchmark UniDAformer over the multi-branch panoptic segmentation architecture ($i.e.$, PSN~\cite{kirillov2019panoptic}). Since HMC introduces little extra computation overhead and can be incorporated as a plug-in, we directly apply HMC (with the online self-training loss) on the multi-branch architecture for benchmarking. Table~\ref{dis:cvrn} shows experimental results on SYNTHIA $\rightarrow$ Cityscapes. We can see that UniDAformer outperforms CVRN in mSQ, mRQ and mPQ consistently. In addition, it similarly improves mRQ by large margins, which further verifies the motivation and design of the proposed HMC.

\renewcommand\arraystretch{1.}
\begin{table}[t]
\centering
\begin{footnotesize}
\resizebox{1.0\linewidth}{!}{
\begin{tabular}{c|ccc|ccc|ccc}
\toprule
\multicolumn{1}{c|}{\multirow{2}{*}{Order}} &\multicolumn{3}{c|}{R$\rightarrow$S$\rightarrow$P}  &\multicolumn{3}{c|}{P$\rightarrow$S$\rightarrow$R}  &\multicolumn{3}{c}{S$\rightarrow$P$\rightarrow$R} \\
\cmidrule{2-10}
&mSQ &mRQ &mPQ &mSQ &mRQ &mPQ &mSQ &mRQ &mPQ \\
\midrule
\multicolumn{1}{c|}{Results}&64.7 &42.2 &33.0  & 62.0 &38.9& 29.9 &62.2  &39.9  &30.1\\
\bottomrule
\end{tabular}
}
\end{footnotesize}
\caption{
The calibration order affects domain adaptation performance. The experiments are conducted over task SYNTHIA $\rightarrow$ Cityscapes. R, S and P denote region-wise calibration, superpixel-wise calibration and pixel-wise calibration, respectively.
}
\label{dis:order}
\end{table}

\renewcommand\arraystretch{1.}
\begin{table}[t]
\centering
\begin{footnotesize}
\resizebox{1.0\linewidth}{!}{
\begin{tabular}{c|c|c|c|c}
\toprule
Methods& Architecture &Parameter & Training Speed & Inference Speed\\
\midrule
CVRN~\cite{huang2021cvrn}& Multi-branch&185.58 M&0.27 fps&0.36 fps \\
UniDAformer&Unified &77.68 M   &1.53 fps  &5.23 fps\\
\bottomrule
\end{tabular}
}
\end{footnotesize}
\caption{Efficiency comparison with multi-branch panoptic adaptation network CVRN~\cite{huang2021cvrn} in terms of parameter number, training speed and inference speed.}
\label{dis:cvrn_p}
\end{table}

\subsection{Discussions}
\label{sec:dis}

\noindent \textbf{Prediction Quality Analysis.} UniDAformer suppresses false predictions effectively via HMC. We examine it over task SYNTHIA $\rightarrow$ Cityscapes with DETR~\cite{carion2020detr}. As discussed in Section~\ref{sec:details}, the predictions in panoptic segmentation consists of three parts including TP, FP and FN. We compute the proportion of each part over all predictions and Fig.~\ref{fig:tp} shows experimental results. We can observe that UniDAformer produces clearly more TP and less FN and FP as compared with both baseline ~\cite{carion2020detr} and the state-of-the-art~\cite{guan2021scale,huang2021cvrn}. This demonstrates the superiority of UniDAformer in suppressing false predictions in domain adaptive panoptic segmentation.

\noindent \textbf{The Calibration Order Matters.} The proposed HMC calibrates predicted pseudo masks in a coarse-to-fine manner ($i.e.$, from region level to superpixel and pixel levels). We study how calibration order affects panoptic segmentation by testing two reversed calibration orders as shown in Table~\ref{dis:order}. It can be seen that reversing calibration order leads to clear performance drops, indicating the benefits of the coarse-to-fine calibration in our design.

\noindent \textbf{Efficiency Comparison with CVRN~\cite{huang2021cvrn}.} Beyond segmentation accuracy, we also benchmark UniDAformer with multi-branch panoptic adaptation network CVRN~\cite{huang2021cvrn} in parameter number, training speed and inference speed. As Table~\ref{dis:cvrn_p} shows, UniDAformer has clearly less parameters and its training and inference time is much shorter than CVRN as well, demonstrating its great simplicity and efficiency.

\section{Conclusion}
This paper presents UniDAformer, a unified domain adaptive panoptic segmentation transformer.
UniDAformer introduces a Hierarchical Mask Calibration (HMC) technique to 
calibrate the predicted pseudo masks on the fly during re-training.
UniDAformer has three unique features: 1) it achieves unified panoptic
adaptation by treating things and stuff as masks and adapting them uniformly; 2) it mitigates the severe false prediction issue effectively by calibrating the predicted pseudo masks iteratively and progressively; 3) it is end-to-end trainable with much less parameters and simpler training and inference pipeline. Besides, the proposed HMC introduces little extra computation overhead and could be used as a plug-in.
Extensive experiments over multiple public benchmarks show that UniDAformer achieves superior segmentation accuracy and efficiency as compared with the state-of-the-art.
Moving forwards, we plan to continue to investigate simple yet effective techniques for unified domain adaptive panoptic segmentation.

\textbf{Acknowledgement.}
This study is supported under the RIE2020 Industry Alignment Fund – Industry Collaboration Projects (IAF-ICP) Funding Initiative, as well as cash and in-kind contribution from the industry partner(s).

{\small
\bibliographystyle{ieee_fullname}
\bibliography{egbib}
}

\end{document}